%% file: main.tex
\documentclass{article}

    \PassOptionsToPackage{numbers, compress, sort}{natbib}
 \usepackage[preprint]{neurips_2026}

\usepackage[utf8]{inputenc} 
\usepackage[T1]{fontenc}    
\usepackage{hyperref}       
\usepackage{url}            
\usepackage{booktabs}       
\usepackage{amsfonts}       
\usepackage{nicefrac}       
\usepackage{microtype}      
\usepackage{xcolor}         

\title{Multi-Modal Multi-Agent Reinforcement Learning\\for Radiology Report Generation%
}

\author{%
  Kaito Baba
  \And
  Risa Kishikawa
  \And
  Satoshi Kodera
  \And
  \\[-9mm]
  Department of Cardiovascular Medicine\\
  The University of Tokyo Hospital, Tokyo, Japan\\
  \texttt{baba-kaito662@g.ecc.u-tokyo.ac.jp}\\[-2mm]
}

\usepackage{amsmath,amssymb}
\usepackage{mathtools}
\usepackage{xspace}
\usepackage[capitalise]{cleveref}
\usepackage{booktabs}
\usepackage{microtype}
\usepackage{tikz}
\usepackage{multirow}
\usepackage{wrapfig}

\makeatletter
  \DeclareRobustCommand\onedot{\futurelet\@let@token\@onedot}
  \def\@onedot{\ifx\@let@token.\else.\null\fi\xspace}
\makeatother
\newcommand\eg{e.g\onedot}

\renewcommand\paragraph[1]{\noindent\textbf{#1}}
\newcommand\best[1]{\textbf{#1}}
\newcommand\worst[1]{\underline{#1}}

\makeatletter
\renewcommand{\normalsize}{%
  \@setfontsize\normalsize\@xpt\@xipt
  \abovedisplayskip      6\p@ \@plus 1.5\p@ \@minus 4\p@
  \abovedisplayshortskip \z@ \@plus 2\p@
  \belowdisplayskip      \abovedisplayskip
  \belowdisplayshortskip 3\p@ \@plus 2\p@ \@minus 2\p@
}
\normalsize
\makeatother

\begin{document}

\maketitle

\input{sec/abs}

\input{sec/intro}

\input{sec/related_work}

\input{sec/method}

\input{sec/experiments}

\input{sec/human_evaluation}

\input{sec/conclusion}



\bibliographystyle{ieeenat_fullname}
\bibliography{main}


\appendix

\crefalias{section}{appendix}
\crefalias{subsection}{appendix}
\crefalias{subsubsection}{appendix}

\input{apx/extended_related_work}

\input{apx/experimental_setup}



\end{document}

%% file: sec/abs.tex
\begin{abstract}
  We propose \textbf{MARL-Rad}, a multi-modal multi-agent reinforcement learning framework for radiology report generation that trains the entire agentic system on policy within its deployed radiology workflow.
  MARL-Rad addresses the limitation of post-hoc agentization, where fixed LLMs are organized into hand-designed agentic workflows without being optimized for their assigned roles.
  Our framework decomposes chest X-ray interpretation into region-specific agents and a global integrating agent, and jointly optimizes them using clinically verifiable rewards.
  Experiments on the MIMIC-CXR and IU X-ray datasets show that MARL-Rad consistently improves clinical efficacy metrics such as RadGraph, CheXbert, and GREEN scores, achieving state-of-the-art clinical efficacy performance.
  Further analyses show that MARL-Rad improves laterality consistency and produces more accurate and detailed reports.
  A blinded clinician evaluation further suggests that MARL-Rad produces reports clinically comparable to ground-truth reports.
\end{abstract}

%% file: sec/intro.tex
\vspace{-3pt}
\section{Introduction}
\vspace{-1mm}

Recent advances in large language models (LLMs) and large vision--language models (LVLMs) have demonstrated remarkable reasoning and generation capabilities across a wide range of domains, including dialogue systems, mathematical reasoning, code synthesis, scientific discovery, and medical diagnosis~\citep{guo2025deepseek,openai2024gpt4technicalreport,comanici2025gemini25pushingfrontier,yang2025qwen3technicalreport}.
To further translate these capabilities into complex real-world tasks, agentic systems have emerged as an increasingly important paradigm for leveraging LLMs beyond single-turn generation.
By decomposing tasks into manageable subtasks, agentic systems can support structured reasoning and workflow-oriented problem solving~\citep{%
  wu2024autogen,%
  hong2024metagpt,%
  hu2025owl,%
  li2023camel,%
  zhao2025llmbasedagenticreasoningframeworks,%
  baba2025proveragentagentbasedframework,%
  plaat2025agenticlargelanguagemodels,%
  koubaa2025pretrainedlanguageagentic,%
  jiang2025largeaimodelsagentic,%
  Li2024surveyllmbasedmultiagentsystems%
}.

\vspace{-1pt}
Medicine is one of the most important application domains of LLMs and LVLMs.
Among medical modalities, chest X-ray (CXR) is one of the most widely used diagnostic tools in clinical practice, enabling physicians to assess thoracic structures and identify conditions such as pneumonia, pneumothorax, and pleural effusion~\citep{Metlay2019CAPGuideline,BRODER2011185,Candemir2019,Alapat2022PneumoniaCXR}.
Radiologists carefully inspect CXR images and manually compose diagnostic reports based on their interpretations.
However, radiology reporting is time-consuming and labor-intensive, and the growing demand for diagnostic imaging can delay diagnosis and degrade report quality~\citep{Cowan2013RadiologistWorkload,Boland2008,Shah2022DelayedTAT,Dong2025}.
To alleviate this burden, automated radiology report generation (RRG) using LLMs has attracted growing attention as a promising approach~\citep{%
  Liu2025EnhancedContrastive,
  Park_2025_CVPR,
  zhou2024largemodeldrivenradiology,
  yan-etal-2023-style,
  lee2025clarifidimprovingradiologyreport,
  Ahsan2025ARDGen,
  wang2024expertinsightenhanced,
  delbrouck-etal-2025-automated,
  LI2025100171,
  castro2025padchestgr,
  Wang2025CXPMRGBench,
  Sirbu2025GITCXR,
  baba2024jradievojapaneseradiologyreport,
  xiao-etal-2025-online
}, with recent studies increasingly exploring agentic-system designs to
improve performance \citep{%
  zhang2025radagentsmultimodalagenticreasoning,
  lou2025cxragentdirectororchestratedmultistagereasoning,
  elboardy2025medicalaiconsensusmultiagent,
  yi2025multimodalmultiagentframeworkradiology,
  kim2024mdagents
}.

\vspace{-1pt}
Despite these recent developments, most existing agentic systems do not train the underlying LLMs for their deployed workflows.
They typically keep pretrained or domain-adapted LLMs fixed and rely on prompts and hand-designed workflows to elicit role-specific behavior~\citep{zhao2025llmbasedagenticreasoningframeworks,plaat2025agenticlargelanguagemodels,koubaa2025pretrainedlanguageagentic,jiang2025largeaimodelsagentic,Li2024surveyllmbasedmultiagentsystems}.
As a result, role specialization and coordination are imposed only through prompt context: the workflow expects agents to act as specialized and coordinated components, while the underlying LLMs remain fixed policies that have not been optimized for their positions in the workflow.
This mismatch is particularly consequential in real-world tasks that require specialized expertise, such as medicine.
In such settings, domain-specific fine-tuning alone is insufficient;
the full agent system must be trained on policy within the deployed workflow.
Indeed, in \cref{sec:ablation_study}, we show that even MedGemma~\citep{sellergren2025medgemmatechnicalreport}, a medical adaptation of Gemma~3~\citep{gemmateam2025gemma3technicalreport}, performs poorly when its parameters are kept fixed and it is merely organized into a multi-agent workflow for RRG.

\vspace{-1pt}
To address this limitation, we propose \textbf{MARL-Rad}, a multi-modal multi-agent reinforcement learning framework that trains the entire agentic RRG system end-to-end on policy within its deployed workflow.
MARL-Rad consists of region-specific agents responsible for localized observations and a global integrating agent that synthesizes their outputs.
These agents are jointly trained with clinically verifiable rewards, optimizing the agentic system within its deployed workflow.
This workflow mirrors the real practice of radiologists, who meticulously examine each anatomical region before composing a comprehensive diagnostic report.
Experiments on the MIMIC-CXR and IU X-ray datasets demonstrate that MARL-Rad consistently improves clinical efficacy (CE) metrics such as RadGraph F1~\citep{jain2021radgraph}, CheXbert F1~\citep{smit-etal-2020-combining}, and GREEN scores~\citep{ostmeier-etal-2024-green}, achieving state-of-the-art performance.
Moreover, deeper analyses show that MARL-Rad improves laterality consistency and produces more detailed and clinically accurate descriptions.
Finally, a small blinded clinician evaluation suggests that MARL-Rad produces clinically comparable reports to ground-truth reports.

Our key contributions are summarized as follows:

\vspace{-2mm}
\begin{itemize}
  \item \textbf{Workflow-aligned training beyond fixed agentification}:
We show that simply organizing a fixed LLM into an agentic workflow is insufficient and can even degrade performance.
MARL-Rad addresses this limitation by training the entire agentic system on policy within the same radiology-inspired workflow in which it is deployed.
  \vspace{-0.7mm}
  \item \textbf{State-of-the-art performance on CE metrics}: Experiments on the MIMIC-CXR and IU X-ray datasets demonstrate that MARL-Rad achieves state-of-the-art performance on various CE metrics, including RadGraph F1, CheXbert F1, and GREEN scores.
  \vspace{-0.7mm}
  \item \textbf{Enhanced laterality consistency and accurate, detail-informed reports}: MARL-Rad improves laterality consistency and generates more detailed and clinically accurate reports compared to single-agent RL baselines.
\end{itemize}

%% file: sec/related_work.tex
\vspace{-2mm}
\section{Related work}
\label{sec:related_work}
\vspace{-1.3mm}

Here we review the prior work most relevant to ours; further discussion is provided in \cref{apx:extended-related-work}.

\vspace{-1pt}
\paragraph{LLMs for radiology report generation (RRG).}
Early RRG systems mostly followed encoder--decoder paradigms with Transformers, such as R2Gen~\citep{chen-etal-2020-generating}, which established strong baselines on the MIMIC-CXR~\citep{mimic-cxr} and IU X-ray~\citep{iu-xray} datasets.
Recent LLM-centric approaches align a medical visual encoder with a frozen or fine-tuned LLM to produce more fluent, clinically grounded reports, such as R2GenGPT~\citep{WANG2023100033}, XrayGPT~\citep{thawakar-etal-2024-xraygpt}, MAIRA-1~\citep{hyland2024maira1specialisedlargemultimodal}, MAIRA-2~\citep{bannur2024maira2groundedradiologyreport}, and CheXagent~\citep{chen2024chexagent}, along with other approaches~\citep{%
  Liu2025ageneralistmedicallanguagemodel,
  pellegrini2025radialoglargevisionlanguagemodel,
  pmlr-v235-yang24v,
  Zhang2024Traditional,
  you2023cxrclip,
  huang2021gloria,
  lee2025cxrllava,
  jiang2025advancingmedical,
  codella2024medimageinsightopensourceembeddingmodel,
  DBLP:conf/iclr/0010WSIRA24,
  ma2024eyegaze,
  moutakanni2024advancinghumancentricairobust,
  Chaves2025LLaVARad,
  Dong2025,
  doi:10.1148/ryai.240790,
  zhou2024largemodeldrivenradiology,
  yan-etal-2023-style,%
  Liu2025EnhancedContrastive,%
  Park_2025_CVPR,%
  lee2025clarifidimprovingradiologyreport,%
  Ahsan2025ARDGen,%
  wang2024expertinsightenhanced,%
  delbrouck-etal-2025-automated,%
  Sirbu2025GITCXR,%
  Heiman_2025_CVPR,%
  Nath_2025_CVPR,%
  Albastaki_2025_CVPR,%
  11147578,%
  kalisch2025ctgraphhierarchicalgraphattention,%
  Hou_2025_ICCV,%
  wang2025curv,%
  DBLP:conf/aaai/Xiao0L0B25,%
  DBLP:conf/aaai/LiuWH0YCGY25,%
  Wang_Chen_Le_Xu_Xu_Zhang_Yang_2025,%
  Zhang_Shi_Ji_Zheng_Qu_2025,%
  Wang_Zhang_Wang_Wang_Hu_Guo_Zhou_Pang_Wen_2025,%
  Huang_Chen_Liu_Lu_Luo_Shen_2025,%
  hou-etal-2025-radar,%
  zhang-etal-2025-libra,%
  xiao-etal-2025-online,%
  kim-etal-2025-look,%
  10.1145/3746027.3754913,%
  10.1145/3746027.3755368,%
  ijcai2025p824,%
  10887699,%
  yin-etal-2025-kia,%
  huang-etal-2025-cmeaa%
}.

\vspace{-1pt}
\paragraph{Agentic systems for RRG.}
Multi-agent frameworks are beginning to appear in RRG, where they align with clinical reasoning stages or combine retrieval-augmented generation (RAG).
RadAgents~\citep{zhang2025radagentsmultimodalagenticreasoning} proposes a radiologist-like, multi-agent workflow for chest X-ray interpretation.
CXRAgent~\citep{lou2025cxragentdirectororchestratedmultistagereasoning} introduces a director-orchestrated, multi-stage agent for chest X-ray interpretation that validates tool outputs with an evidence-driven validator and coordinates diagnostic planning and team-based reasoning.
\citet{yi2025multimodalmultiagentframeworkradiology} decomposes CXR reporting into retrieval, draft, refinement, vision, and synthesis agents aligned with stepwise clinical reasoning.
\citet{elboardy2025medicalaiconsensusmultiagent} proposes a model-agnostic, ten-agent framework that unifies radiology report generation and evaluation, coordinated by an orchestrator and including an LLM-as-a-judge.
However, these agentic RRG systems are mostly training-free, relying on pretrained models without end-to-end optimization of entire systems.

\vspace{-1pt}
\paragraph{Reinforcement learning for RRG.}
Recent studies have incorporated reinforcement learning (RL) to enhance clinical accuracy in RRG.
LM-RRG~\citep{zhou2024largemodeldrivenradiology} integrates clinical-quality RL by directly optimizing RadCliQ~\citep{Yu2023EvaluatingProgressRRG} as a reward signal.
BoxMed-RL~\citep{jing2025reasonlikeradiologistchainofthought} couples chain-of-thought supervision with spatially verifiable RL that ties textual findings to bounding-box evidence.
DeepMedix-R1~\citep{lin2025foundationmodelchestxray} employs a three-stage pipeline: instruction-fine-tuning, synthetic reasoning sample exposure, and online RL.
OraPO~\citep{chen2025orapooracleeducatedreinforcementlearning} proposes an oracle-educated group relative policy optimization (GRPO) for RRG that leverages fact-level rewards.
Med-R1~\citep{lai2025medr1reinforcementlearninggeneralizable} applies GRPO-based RL to medical VLMs across eight imaging modalities and five VQA task types.
However, these RL approaches primarily focus on optimizing a single model rather than optimizing entire multi-agent systems.

%% file: sec/method.tex
\section{Method}

\subsection{Preliminaries}

In this section, we introduce the notations necessary for this paper, and as background, briefly describe Group Sequence Policy Optimization (GSPO)~\citep{zheng2025groupsequencepolicyoptimization}, which is a sequence-level variant of GRPO~\citep{shao2024deepseekmathpushinglimitsmathematical} that improved performance.

Let $\mathcal{D}$ denote the data distribution over query--answer pairs $(q,a)$.
Like GRPO, GSPO samples a group of $G$ responses
$
  (\{x_i\}_{i=1}^G \sim \pi_{\theta_{\text{old}}}(\cdot\mid q))
$
for each query $q$,
where $\pi_{\theta_{\text{old}}}$ denotes the policy used to generate outputs before updating the current policy.
Each sampled response $x_i$ is then evaluated by a verifiable reward $r(x_i,a)$, and group-relative advantage is computed as follows:
\begin{equation}
  \hat{A}_i
  \coloneqq \frac{
      r(x_i, a)-\operatorname{mean}(\{r(x_i,a)\}_{i=1}^G)
    }{
      \operatorname{std}(\{r(x_i,a)\}_{i=1}^G)
    }.
  \label{eq:group_relative_advantage}
\end{equation}
GSPO then performs policy optimization by maximizing the following objective:
\begin{align}
  \!\!\!\mathcal{J}_{\text{GSPO}}(\theta)
  \!\coloneqq\!
  \mathbb{E}_{
    (q,a)\sim\mathcal{D},
    \{x_i\}_{i=1}^G\!\sim \pi_{\theta_{\text{old}}}\!(\cdot\mid q)
  }\!
  \Bigg[
    \!\frac{1}{G}\!\sum_{i=1}^G
    \min\! \left(
      \!s_i(\theta)\hat{A}_i,
      \operatorname{clip}(
        s_i(\theta),
        1-\varepsilon_\mathrm{low},
        1+\varepsilon_\mathrm{high}
      )
      \hat{A}_i\!
    \right)
  \!\!\Bigg]\!,\!\!
\end{align}
where
\begin{align}
  s_i(\theta)
  &\coloneqq
  \left(
    \frac{
      \pi_\theta(x_i\mid q)
    }{
      \pi_{\theta_{\text{old}}
    }(x_i\mid q)}
  \right)^{\!1/|x_i|}
  =
  \left(
    \frac{
      \prod_{t=1}^{|x_i|}
      \pi_\theta(x_{i,t}\mid q,x_{i,<t})
    }{
      \prod_{t=1}^{|x_i|}
      \pi_{\theta_\text{old}}(x_{i,t}\mid q,x_{i,<t})
    }
  \right)^{\!1/|x_i|}
\end{align}
is the importance ratio based on sequence likelihoods,
and $|x_i|$ denotes the number of tokens of response $x_i$.
Here, $x_{i,t}$ and $x_{i,<t}\coloneqq (x_{i,1},\ldots,x_{i,t-1})$ denote the $t$-th tokens and the preceding tokens of the response $x_i$, respectively.

\subsection{Multi-agent GSPO}
\label{sec:ma_gspo}

In this section, we introduce a simple yet effective multi-agent formulation of GSPO.
Although the formulation is a simple extension of GSPO, it enables the entire agentic system to be optimized end-to-end in an on-policy manner, encouraging coordination among agents within the deployed workflow.
The adaptation to RRG is discussed in \cref{sec:rrg_method}.

Let there be $K$ agents with policies $\{\pi_{\theta^{(k)}}\}_{k=1}^K$,
where the indices $k=1,\ldots,K$ follow the order in which the agents are activated.
For each $(q,a)\!\sim\!\mathcal{D}$, we sample a \emph{group of $G$ joint rollouts}:
\begin{align}
  \{x_i\}_{i=1}^G
  \sim
  \pi_{\theta_\text{old}}^\text{agent}(\cdot\mid q)
  \coloneqq
  \prod_{k=1}^{K}
  \pi_{\theta_\text{old}^{(k)}}(\,\cdot\,\mid c^{(k)}_i),
\end{align}
where $\smash{x_i \coloneqq (x^{(1)}_i,\ldots,x^{(K)}_i)}$. Here, $\smash{c^{(k)}_i}$ denotes the context observed by agent $k$ when generating
$x^{(k)}_i$. This context may include the query $q$ or the outputs of preceding agents.

Each joint rollout is evaluated by a system-level verifiable reward
$
\smash{r_i \coloneqq r(x_i^{(K)},a),}
$
and the resulting group-relative advantage computed by \cref{eq:group_relative_advantage} is shared across all agents.

Then, we define the \emph{multi-agent GSPO (MA-GSPO)} objective as follows:
\begin{align}
  &\!\!\mathcal{J}_{\text{MA-GSPO}}(\theta_1,\ldots,\theta_K)
  \nonumber\\[-3mm]
  &\!\!\!=\!
  \mathbb{E}_{
    (q,a)\sim\mathcal{D},
    \{x_i\}\sim\pi_{\theta_\text{old}}^\text{agent}(\cdot\mid q)
  }\!
  \Bigg[\!
    \frac{1}{K}\!\sum_{k=1}^{K}
    \!\frac{1}{G}\!\sum_{i=1}^{G}
    \min\!\Big(\!
      s^{(k)}_i\!(\theta_k)\hat{A}_i,
      \operatorname{clip}\!\big(s^{(k)}_i\!(\theta_k),1-\varepsilon_\mathrm{low},1+\varepsilon_\mathrm{high}\big)\hat{A}_i\!
    \Big)\!
  \Bigg]\!,\!\!
\end{align}
where the importance ratio for agent $k$ is:

\vspace{-6mm}
\begin{equation}
  s^{(k)}_i(\theta_k)
  =
  \Bigg(
    \frac{
      \pi_{\theta_k}\!\big(x^{(k)}_i \mid c^{(k)}_i\big)
    }{
      \pi_{\theta^{\text{old}}_k}\!\big(x^{(k)}_i \mid c^{(k)}_i\big)
    }
  \Bigg)^{\!1/|x^{(k)}_i|}.
\end{equation}

This reduces to standard GSPO when $K=1$.
Although the advantage is shared, each agent is updated through its own sequence-level importance ratio, so the objective optimizes the joint agentic system while preserving agent-specific policy updates.

\input{figures/workflow}

\subsection{Multi-agent RL on RRG}
\label{sec:rrg_method}
\vspace{-1.5mm}

The overall architecture of MARL-Rad is illustrated in \cref{fig:workflow}.
Our system consists of three region-specific agents and one global integrating agent
that collaborate to generate the final diagnostic report.
Each region-specific agent focuses on a distinct anatomical region of the chest X-ray:
a \emph{left-region agent} examines structures such as the left lung, left hilar structures, left costophrenic angle, and left clavicle;
a \emph{right-region agent} examines structures such as the right lung, right hilar structures, right costophrenic angle, and right clavicle;
and a \emph{central-region agent} examines structures such as the cardiac silhouette, mediastinum, aortic arch, trachea, and spine.
This design follows the actual diagnostic workflow of radiologists,
who meticulously examine each anatomical region of a chest X-ray
before composing the complete report.
The \emph{global integrating agent} then receives the outputs from the three region-specific agents, incorporates their diagnoses, and produces a final comprehensive report reflecting the overall condition of the chest X-ray.

\vspace{-2pt}
For the reinforcement learning, we employ clinically verifiable rewards using the CheXbert~\citep{smit-etal-2020-combining} accuracy, which measures the correctness of predicted clinical findings based on automatically labeled disease categories, and the RadGraph F1~\citep{jain2021radgraph}, which evaluates factual and relational consistency.
Additionally, we incorporate the ROUGE-L~\citep{lin-2004-rouge} to encourage lexical alignment with reference reports.
The final reward is defined as the unweighted sum of these three components, and the entire agent system is jointly optimized as we described in \cref{sec:ma_gspo}.

%% file: figures/workflow.tex
\tikzset{%
  network/.pic = {%
    \node[circle, draw, semithick, inner sep=1.5pt] (a1) at (-0.5, 3mm) {};
    \node[circle, draw, semithick, inner sep=1.5pt] (a2) at (-0.5, -3mm) {};
    \node[circle, draw, semithick, inner sep=1.5pt] (b1) at (0, 4mm) {};
    \node[circle, draw, semithick, inner sep=1.5pt] (b2) at (0, 0mm) {};
    \node[circle, draw, semithick, inner sep=1.5pt] (b3) at (0, -4mm) {};
    \node[circle, draw, semithick, inner sep=1.5pt] (c1) at (0.5, 0) {};
    \draw[semithick] (a1) -- (b1);
    \draw[semithick] (a1) -- (b2);
    \draw[semithick] (a1) -- (b3);
    \draw[semithick] (a2) -- (b1);
    \draw[semithick] (a2) -- (b2);
    \draw[semithick] (a2) -- (b3);
    \draw[semithick] (b1) -- (c1);
    \draw[semithick] (b2) -- (c1);
    \draw[semithick] (b3) -- (c1);
  }%
}

\definecolor{myBlue}{HTML}{0072B2}
\definecolor{skyblue}{HTML}{56B4E9}
\definecolor{myOrange}{HTML}{D55E00}
\definecolor{myPink}{HTML}{CC79A7}
\definecolor{myYellow}{HTML}{F0E442}
\definecolor{myGreen}{HTML}{009E73}

\newcommand{\defineModel}[3]{%
  \tikzset{%
    #1/.pic = {%
      \filldraw[rounded corners, fill=#3, draw=black] (-13mm, -8mm) rectangle (13mm, 8mm);
      \node at (0, -5.5mm) {\fontsize{6.8pt}{6.8pt}\selectfont\textsf{#2}};
      \pic[anchor=south] at (0, 0.5mm) {network};
      \node at (9mm, 6mm) {\includegraphics[width=6mm]{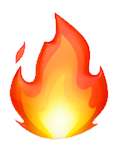}};
    }%
  }%
}

\defineModel{left}{Left-Region Agent}{myYellow!30}
\defineModel{right}{Right-Region Agent}{myOrange!30}
\defineModel{center}{Central-Region Agent}{myPink!30}
\defineModel{global}{Global Integrating Agent}{myBlue!30}

\newcommand{\noteFontPrimary}[1]{\fontsize{9pt}{8pt}\selectfont\textbf{\textsf{\color{myBlue}#1}}}
\newcommand{\noteFontSecondary}[1]{\fontsize{9pt}{8pt}\selectfont\textsf{#1}}
\newcommand{\noteFontPrimaryPink}[1]{\fontsize{9pt}{8pt}\selectfont\textbf{\textsf{\color{myPink}#1}}}

\begin{figure*}[t]
  \centering
  \begin{tikzpicture}[scale=0.51, every node/.style={scale=0.51}]

    \node at (0mm, -18mm) {
      \noteFontPrimary{Chest X-ray}
    };
    \node at (0, 0) {\includegraphics[width=30mm]{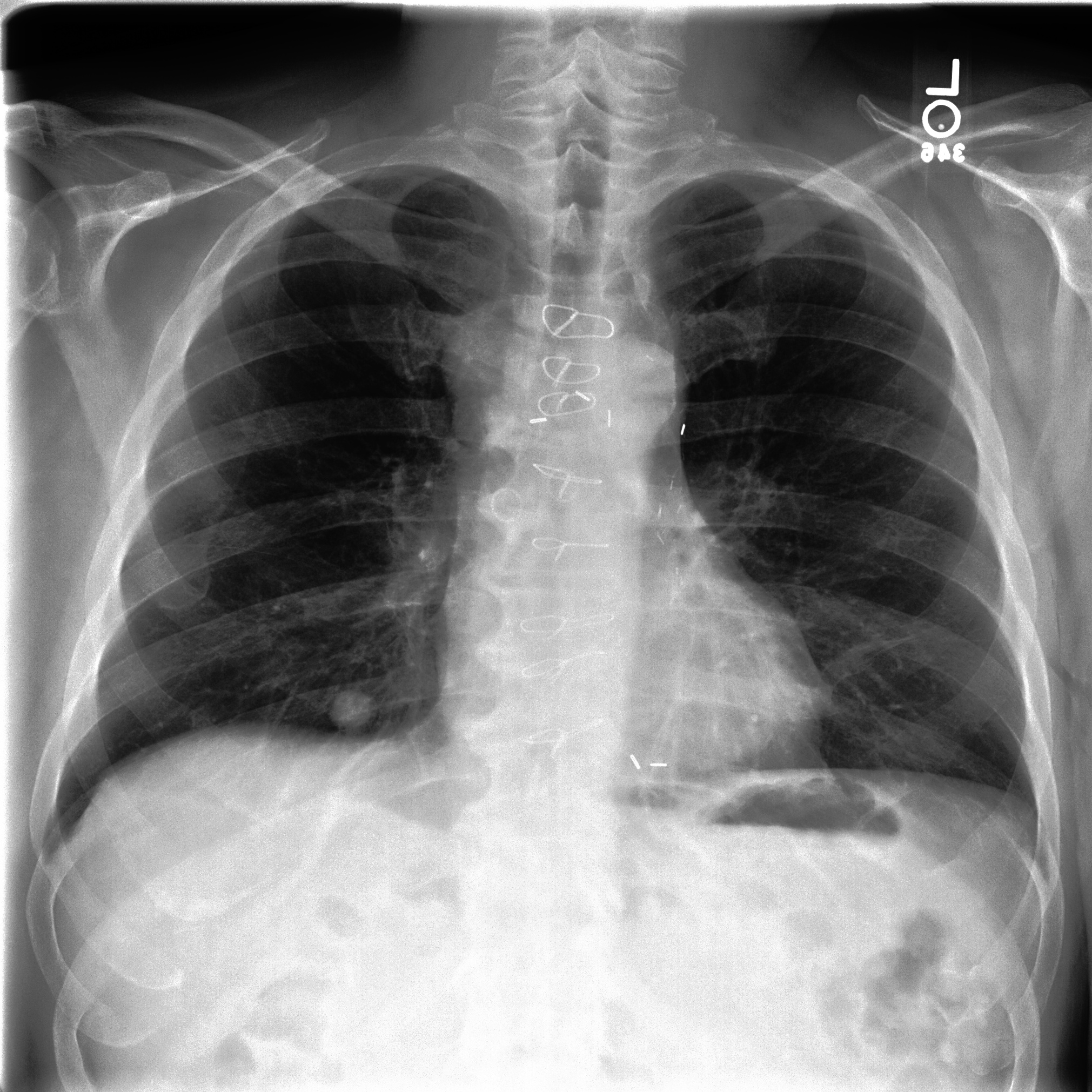}};

    \draw[->, myBlue, very thick] (17mm, 1mm) -- ++(8mm, 17mm) -- ++(9mm, 0);
    \draw[->, myBlue, very thick] (17mm, 0mm) -- ++(17mm, 0);
    \draw[->, myBlue, very thick] (17mm, -1mm) -- ++(8mm, -17mm) -- ++(9mm, 0);

    \pic at (48mm, 18mm) {left};
    \pic at (48mm, 0mm) {center};
    \pic at (48mm, -18mm) {right};

    \draw[->, myBlue, very thick] (62mm, 17mm) -- ++(20mm, 0) -- ++(29mm, -11mm);
    \draw[->, myBlue, very thick] (62mm, 0mm) -- ++(49mm, 0);
    \draw[->, myBlue, very thick] (62mm, -17mm) -- ++(20mm, 0) -- ++(29mm, 11mm);

    \pic at (125mm, 0mm) {global};

    \draw[->, myBlue, very thick] (139mm, 0mm) -- ++(8mm, 0);

    \filldraw[fill=myGreen!30, draw=myGreen!40] (148mm, -10mm) rectangle ++(42mm, 20mm);
    \node[anchor=west, align=left] at (148mm, 0mm) {
      \noteFontSecondary{\#\#\# Findings}\\
      \noteFontSecondary{The lungs are clear ...}\\
      \noteFontSecondary{\#\#\# Impression}\\
      \noteFontSecondary{No acute cardiopulmonary...}
    };
    \node at (169mm, -13mm) {
      \noteFontPrimary{Final Generated Report}
    };

    \draw[<->, myPink, very thick] (191mm, 0mm) -- ++(16mm, 0);

    \filldraw[fill=myGreen!30, draw=myGreen!40] (208mm, -10mm) rectangle ++(42mm, 20mm);
    \node[anchor=west, align=left] at (208mm, 0mm) {
      \noteFontSecondary{\#\#\# Findings}\\
      \noteFontSecondary{The lungs are clear ...}\\
      \noteFontSecondary{\#\#\# Impression}\\
      \noteFontSecondary{No acute cardiopulmonary...}
    };
    \node at (229mm, -13mm) {
      \noteFontPrimary{Ground Truth Report}
    };

    \draw[->, myPink, very thick] (199mm, 1mm) -- ++(0, 26mm) -- ++(-134mm, 0) -- ++(-4mm, -1mm);

    \draw[->, myPink, very thick] (139mm, 27mm) -- ++(-5mm, -17mm);
    \draw[->, myPink, very thick] (139mm, 27mm) -- ++(-78mm, -19mm);
    \draw[->, myPink, very thick] (139mm, 27mm) -- ++(-78mm, -38mm);

    \node at (175mm, 24mm) {
      \noteFontPrimaryPink{Clinically Verifiable Reward}
    };

    \node[anchor=west, align=left] at (61mm, 21mm) {
      \noteFontSecondary{Examines left lung,}\\[-2pt]
      \noteFontSecondary{left hilar structures, ...}
    };
    \node[anchor=west, align=left] at (61mm, 4mm) {
      \noteFontSecondary{Examines cardiac silhouette,}\\[-2pt]
      \noteFontSecondary{mediastinum, trachea, ...}
    };
    \node[anchor=west, align=left] at (61mm, -21mm) {
      \noteFontSecondary{Examines right lung,}\\[-2pt]
      \noteFontSecondary{right hilar structures, ...}
    };

    \node[anchor=west, align=left] at (97mm, 13mm) {
      \noteFontPrimary{Diagnosis of}\\[-3pt]
      \noteFontPrimary{\hspace{3mm}left region}
    };
    \node[anchor=west, align=left] at (85mm, -4mm) {
      \noteFontPrimary{Diagnosis of}\\[-3pt]
      \noteFontPrimary{central region}
    };
    \node[anchor=west, align=left] at (97mm, -13mm) {
      \noteFontPrimary{\hspace{3mm}Diagnosis of}\\[-3pt]
      \noteFontPrimary{right region}
    };

  \end{tikzpicture}
  \vspace{-3pt}
  \caption{
    Overview of the proposed multi-agent RL framework.
    Region-specific agents and the global integrating agent collaboratively generate the radiology report, and the entire agent system is jointly optimized through RL based on clinically verifiable rewards.
  }
  \label{fig:workflow}
\end{figure*}

%% file: sec/experiments.tex
\vspace{-2mm}
\section{Experiments}
\label{sec:experiments}
\vspace{-2mm}
\subsection{Experimental setup}
\label{sec:experimental_setup}
\vspace{-1.5mm}

In this section, we describe the overview of the experimental setup.
Further details are provided in \cref{appx:experimental_setup}.

\paragraph{Datasets}
We train our agents using the MIMIC-CXR~\citep{mimic-cxr} dataset.
Following prior studies~\citep{%
  chen-etal-2020-generating,%
  lee2025clarifidimprovingradiologyreport,%
  baba2024jradievojapaneseradiologyreport,%
  10203622,%
  NICOLSON2024101585,%
  10.1145/3664647.3680760%
}, we exclude samples without the corresponding reports.
We adopt the official split of MIMIC-CXR and use the training set for training our agents.
Evaluation is performed on the official test split of MIMIC-CXR and the IU X-ray~\citep{iu-xray} dataset.
For IU X-ray, we follow the standard 70\%/20\%/10\% split used in prior work~\citep{%
  chen-etal-2020-generating,%
  chen-etal-2021-cross-modal,%
  liu-etal-2021-competence,%
  Hou_2025_ICCV,%
  DBLP:conf/aaai/Xiao0L0B25,%
  chen-etal-2021-cross-modal,%
  Huang_Chen_Liu_Lu_Luo_Shen_2025,%
  Shen_Pei_Liu_Tian_2024,%
  10203079,%
  10203622%
} and use only the test split for cross-dataset evaluation.

\input{tables/mimic_cxr_results}

\paragraph{Evaluation metrics}
Following previous studies~\citep{%
chen-etal-2020-generating,%
chen-etal-2021-cross-modal,%
yan-etal-2023-style,%
10204026,%
10203079,%
10203622,%
10.1007/978-3-031-72775-7_10,%
Shen_Pei_Liu_Tian_2024,%
10.1145/3664647.3680760,%
Liu_Structural_MICCAI2024,%
10356722,%
10.1007/978-3-031-73001-6_11,%
chen-etal-2021-cross-modal,%
NICOLSON2024101585,%
Liu_Tian_Chen_Song_Zhang_2024,%
Liu2025EnhancedContrastive,%
Park_2025_CVPR,%
zhou2024largemodeldrivenradiology,%
lee2025clarifidimprovingradiologyreport,%
Ahsan2025ARDGen,%
lin2025foundationmodelchestxray,%
zhang2025radagentsmultimodalagenticreasoning,%
Hou_2025_ICCV,%
DBLP:conf/aaai/Xiao0L0B25,%
Huang_Chen_Liu_Lu_Luo_Shen_2025,%
xiao-etal-2025-online,%
wang2025interpretable%
}, we evaluate the generated reports using both natural language generation (NLG) metrics and CE metrics.
For NLG metrics, we report BLEU-1, BLEU-4~\citep{papineni-etal-2002-bleu}, METEOR~\citep{banerjee-lavie-2005-meteor}, and ROUGE-L~\citep{lin-2004-rouge}.
For CE metrics, we report RadGraph F1~\citep{jain2021radgraph}, CheXbert F1~\citep{smit-etal-2020-combining}, and GREEN scores~\citep{ostmeier-etal-2024-green}.
CheXbert is a 14-label classification models that assess the presence of common thoracic findings (\eg, pneumonia, atelectasis, edema) from generated reports.
RadGraph measures the correctness of clinical entity and relation extraction, reflecting the structural and factual consistency of the report.
The GREEN score is a LLM-based metric that evaluates clinical efficacy by grading report-level clinical correctness beyond surface-level lexical overlap.
Consistent with prior work, evaluations are performed on the Findings section alone~\citep{%
  Hou_2025_ICCV,%
  Liu2025EnhancedContrastive,%
  chen-etal-2021-cross-modal,%
  lee2025clarifidimprovingradiologyreport,%
  NICOLSON2024101585%
} and on both the Findings and Impression sections combined~\citep{%
  Liu2025EnhancedContrastive,%
  Tanno2025collaborationcliniciansvision,%
  lee2025clarifidimprovingradiologyreport%
}.

\paragraph{Used models}
We use MedGemma-4B~\citep{sellergren2025medgemmatechnicalreport} as the base model for all agents,
including the three region-specific agents and the global integrating agent.
All agents are trained jointly end-to-end from the same base checkpoint, but they do not share parameters, yielding distinct parameter sets.

\input{tables/iu_xray_results}

\vspace{-1mm}
\subsection{Main results: Comparison with previous state-of-the-art methods}
\vspace{-0.6mm}

The results are shown in \cref{tab:mimic_cxr_results,tab:iu_xray_results}.
Our method outperforms the baselines on all CE metrics.
As demonstrated in \citet{Tanno2025collaborationcliniciansvision}, where evaluations were conducted by 27 board-certified radiologists from two regions (the United States and India), traditional NLG metrics based solely on linguistic similarity do not align with clinicians' assessments, whereas CE metrics are more consistent with clinical judgments.
Multiple studies have similarly reported that conventional NLG metrics fail to reflect clinicians' evaluations, underscoring the need to prioritize CE metrics~\citep{pmlr-v106-liu19a,pmlr-v116-boag20a,Yu2023EvaluatingProgressRRG}.
Considering this, it is reasonable that our method does not align with some NLG metrics, while the strong performance on CE metrics suggests improved clinical relevance.

The consistent improvement on both the MIMIC-CXR and IU X-ray datasets demonstrates the effectiveness across different evaluation settings.
Notably, despite being trained exclusively on MIMIC-CXR training set, our agent attains high performance on IU X-ray, suggesting promising robustness and cross-dataset generalization.

We also report the VRAM usage and inference throughput of MARL-Rad in \cref{appx:computational_cost}, showing that despite the additional computation introduced by agentization, it maintains a practical inference speed for real-world radiology reporting workflows.

\subsection{Ablation study}
\label{sec:ablation_study}
\vspace{-3pt}

We conduct ablation studies to examine the effects of workflow-aligned RL, regional decomposition, and reward design.
The results are presented in \cref{tab:ablation_study}.

\vspace{-4pt}
\subsubsection{Training-free agentic workflow is insufficient}
\vspace{-4pt}

First, we evaluate a training-free agentic workflow in which MedGemma is kept fixed and executed with the same agentization as MARL-Rad.
The results are presented as ``MedGemma (agent)'' in \cref{tab:ablation_study}.
This variant performs worse than the vanilla MedGemma model, despite following the same multi-agent workflow as our full method.
This suggests that simply organizing a fixed LLM into an agentic workflow is insufficient and can even degrade performance, likely because the underlying model has not been optimized for its assigned roles or for coordinated reasoning within the deployed workflow.
This finding highlights the importance of jointly optimizing the entire agent system, rather than relying on naive agentification without learning.

\vspace{-3pt}
Furthermore, our method also achieves higher scores than GSPO training without agentization (presented as ``MedGemma + RL''), indicating that our regional multi-agent workflow itself is effective when the agent system is properly optimized for the deployed workflow.
We conduct a more in-depth analysis of this advantage in \cref{sec:region_analysis}.

\input{tables/ablation_study}

\vspace{-3pt}
\subsubsection{Shared versus counterfactual rewards}
\vspace{-4pt}

A natural concern with using a shared system-level reward is that it may suffer from a credit assignment problem: since all agents receive the same final reward, it may be unclear which agent contributed to the improvement or degradation of the final report.
To examine this issue, we consider a naive credit-assignment variant based on counterfactual rewards.
Let $r_i$ denote the reward of the original final report for the $i$-th joint rollout.
For a regional agent $k$, let $\smash{r_i^{(\setminus k)}}$ denote the reward obtained after regenerating the final report with the global integrating agent while removing agent $k$'s intermediate output from its input context.
We then assign agent $k$ the difference reward
$
  \smash{r_i^{(k)}
  \coloneqq
  r_i - r_i^{(\setminus k)}}
$.
That is, each regional agent is rewarded according to how much the final report reward decreases when its regional output is removed from the integrating agent's context.
For the global integrating agent, we keep the original system-level reward $r_i$.
We report this variant as ``Counterfactual Reward'' in \cref{tab:ablation_study}.

As shown in \cref{tab:ablation_study}, this counterfactual reward variant does not improve over our shared-reward formulation and instead yields slightly worse clinical metrics.
This suggests that explicitly decomposing the final reward in this naive manner is not necessarily beneficial for RRG.
One reason is that the counterfactual report is generated from an ablated context that differs from the deployed workflow.
Thus, the resulting reward difference reflects not only the contribution of the removed regional agent, but also the global integrating agent's ability to compensate under an out-of-distribution context.
In contrast, the success of our simple shared-reward formulation shows that explicit counterfactual reward decomposition is not necessary for this setting.
For structured short-horizon agentic RRG workflows, optimizing the complete system with a shared clinically grounded reward is sufficient and preferable to noisy counterfactual reward decomposition.

\subsection{Analysis of laterality consistency using RadGraph}
\label{sec:region_analysis}
\input{tables/region_analysis}

To assess the effect of decomposing the task into region-specific agents, we conduct a detailed analysis of the RadGraph outputs to examine region-level consistency.
RadGraph extracts entities classified as either ``Anatomy'' or ``Observation'' along with the relations between them, enabling the identification of clinically meaningful region--finding pairs, such as ``left--lung--pneumonia.''
Among region-level attributes, laterality (left vs.\ right) is the one most directly handled by our left/right agents, and it carries critical clinical importance: errors in left--right identification are strictly unacceptable in real clinical practice.
Therefore, we extract representative laterality-related entities from the RadGraph output such as ``left,'' ``right,'' ``left lung,'' and ``right lung,'' along with the entities connected to them via relations.
Based on these subsets, we define a laterality-specific RadGraph F1 score by applying the standard RadGraph evaluation procedure only to these laterality-focused subgraphs and use it for our analysis.
The detailed procedure for computing the laterality-specific RadGraph F1 score is described in \cref{appx:detail_region_analysis}.

\input{figures/case_study}

The results are shown in \cref{tab:region_analysis}. We observe that MARL-Rad captures laterality more accurately than the single-agent RL baseline.
This improvement may be attributed to a limitation identified in recent studies, which report that standard vision--language models tend to behave as global image parsers and struggle to reason about spatial relations at the region level~\citep{cheng2024spatialrgpt,Chen_2024_CVPR}. By introducing region-specific agents---including dedicated left/right agents---our approach decomposes the interpretation of image regions and enables each agent to focus on its corresponding region, thereby enhancing the model's ability to capture laterality.

To further verify that the improvement is not merely driven by the global integrating agent, we also report the laterality-specific RadGraph F1 scores of the left and right agents in \cref{tab:region_analysis}.
These scores evaluate each regional agent's output before global integration.
As shown in the table, each left/right agent achieves a score comparable to that of the final integrated report on its assigned side, while its score on the opposite side remains much lower.
This indicates that MARL-Rad does not simply rely on the global agent to correct or infer regional information; rather, the regional agents themselves learn to produce clinically meaningful localized diagnoses.

\subsection{Case study}
\label{sec:case_study}

\cref{fig:case_study} shows an example case from our multi-agent RL system. In this example, the left, right, and central region agents each focus on their respective anatomical responsibilities and consistently produce findings and impression texts that are restricted to their assigned regions. This behavior indicates that the intended task decomposition is functioning as designed.
Furthermore, the global integrating agent leverages these region-specific drafts and produces a concise and coherent final report, rather than simply concatenating the regional outputs.
For instance, given the left and right agents' statements ``The left lung is relatively clear'' and ``The right lung is clear,'' the integrating agent appropriately compresses these into ``The lungs are clear.'' Similarly, central findings such as ``The heart size is normal'' are preserved and incorporated into the final report.
In addition, the integrating agent also adds global findings, such as post-surgical changes (``There is evidence of prior median sternotomy with surgical clips noted in the anterior mediastinum''), which are not tied to any specific region.
As a result, the final report accurately captures both the absence of acute pulmonary abnormalities and the patient's status post cardiac surgery, aligning with the ground-truth conclusion of ``No acute cardiopulmonary process.''

\input{figures/comparison_case}

For comparison, \cref{fig:case_study_comparison} presents an example generated by the single-agent RL model on the same image, where MedGemma is optimized with RL without agentization.
Although the content of the report and its final conclusion are broadly consistent with the ground truth, the report lacks detailed analysis and remains somewhat vague, omitting several clinically relevant findings present in the ground-truth report.
This contrast underscores the advantage of agentization, which explicitly examines each anatomical region and yields more accurate, detailed reports.

Additionally, as illustrated in this example, agentization improves interpretability: each regional agent produces an explicit, region-focused diagnosis, making it clear how the model assessed each anatomical region. This transparency is particularly valuable for clinicians or end-users who may not be familiar with the internal behavior of LVLMs. In contrast, a single-agent model provides only a monolithic summary, making it difficult to verify whether the underlying regional findings were adequately considered.

%% file: tables/mimic_cxr_results.tex
\begin{table*}
  \centering
  \setlength{\tabcolsep}{3.6pt}
  \caption{
    Comparison with previous state-of-the-art methods on the MIMIC-CXR dataset~\citep{mimic-cxr}.
    The best score for each metric within each section is highlighted in \textbf{bold}.
    Our approach achieves the best performance across all clinical efficacy (CE) metrics, which are considered to align more closely with clinicians' assessments~\citep{Tanno2025collaborationcliniciansvision,pmlr-v106-liu19a,pmlr-v116-boag20a,Yu2023EvaluatingProgressRRG}.
  }
  \scalebox{0.85}{
    \begin{tabular}{cccccccc}
      \toprule
      \multirow{2}{*}{\raisebox{-1mm}{\textbf{Method}}}
        & \multicolumn{4}{c}{\textbf{NLG Metrics}$\ \uparrow$}
        & \multicolumn{3}{c}{\textbf{CE Metrics}$\ \uparrow$}\\
      \cmidrule(r){2-5}
      \cmidrule(r){6-8}
      &  \textbf{BLEU-1} & \textbf{BLEU-4}& \textbf{METEOR}& \textbf{ROUGE-L} & \textbf{RadGraph F1} & \textbf{CheXbert F1} & \textbf{GREEN}\\
      \midrule
      \midrule
      \multicolumn{8}{c}{\textit{MIMIC-CXR Findings}}\\
      \midrule

      R2Gen~\citep{chen-etal-2020-generating} & 0.353 & 0.103 & 0.142 & 0.277 & -  & 0.276 & -\\
      R2GenCMN~\citep{chen-etal-2021-cross-modal} & 0.353 & 0.106 & 0.142 & 0.278 & - & 0.278 & -\\
      PPKED~\citep{9578840} & 0.360 & 0.106 & 0.149 & 0.284 & - & - & - \\
      CMCL~\citep{liu-etal-2021-competence} & 0.344 & 0.097 & 0.133 & 0.281 & - & - & -\\
      SA~\citep{yan-etal-2023-style} & - & - & - & - & 0.228 & - & -\\
      RGRG~\citep{10204026} & 0.373 & 0.126 & 0.168 & 0.264 & - & 0.447 & - \\
      METransformer~\citep{10203079} & 0.386 & 0.124 & 0.152 & 0.291 & - & 0.311 & - \\
      KiUT~\citep{10203622} & 0.393 & 0.113 & 0.160 & 0.285 & - & 0.321 & -\\
      CoFE~\citep{10.1007/978-3-031-72775-7_10} & - & 0.125 & 0.176 & 0.304 & - & 0.405 & -\\
      MAN~\citep{Shen_Pei_Liu_Tian_2024} & 0.396 & 0.115 & 0.151 & 0.274 & - & 0.389 & -\\
      Med-LMM~\citep{10.1145/3664647.3680760} & - & 0.128 & 0.161 & 0.289 & - & 0.395 & -\\
      SEI~\citep{Liu_Structural_MICCAI2024} & - & 0.135 & 0.158 & 0.299 & 0.249 & 0.460 & -\\
      FMVP~\citep{10356722} & 0.389 & 0.108 & 0.150 & 0.284 & - & 0.336 & -\\
      HERGen~\citep{10.1007/978-3-031-73001-6_11} & 0.395 & 0.122 & 0.156 & 0.285 & - & 0.317 & -\\
      CMN~\citep{chen-etal-2021-cross-modal} & 0.353 & 0.106 & 0.142 & 0.278 & -  & 0.278 & -\\
      CXRMate~\citep{NICOLSON2024101585} & - & 0.079 & - & 0.262 & 0.272 & 0.357 & -\\
      I3+C2FD~\citep{Liu_Tian_Chen_Song_Zhang_2024} & 0.402 & 0.128 & 0.175 & 0.291 & - & 0.473 & -\\
      MLRG~\citep{Liu2025EnhancedContrastive} & 0.411 & 0.158 & 0.176 & 0.320 & 0.291 & 0.505 &  0.353\\
      DART~\citep{Park_2025_CVPR} & 0.437 & 0.137 & 0.175 & 0.310 & - & 0.533 & -\\
      LM-RRG~\citep{zhou2024largemodeldrivenradiology} & - & 0.122 & 0.165 & 0.296 & - & 0.484 & -\\
      DeepMedix-R1~\citep{lin2025foundationmodelchestxray} & 0.340 & 0.105 & 0.289 & \best{0.329} & 0.238 & 0.239 & -\\
      SRRG~\citep{Hou_2025_ICCV} & 0.416 & 0.128 & 0.294 & 0.162 & - & 0.486 & -\\
      MPO~\citep{DBLP:conf/aaai/Xiao0L0B25} & 0.416 & 0.139 & 0.162 & 0.309 & - & 0.353 & -\\
      DAMPER~\citep{Huang_Chen_Liu_Lu_Luo_Shen_2025} & 0.402 & \best{0.193} & 0.289 & 0.301& - & 0.507 & -\\
      OISA~\citep{xiao-etal-2025-online} & 0.428 & 0.129 & - & - &  0.244 & 0.486 & 0.322\\
      MedGemma~\citep{sellergren2025medgemmatechnicalreport} & 0.285 & 0.014 & 0.238 & 0.203 & 0.195 &0.494 & 0.361\\
      MARL-Rad (Ours) & \best{0.533} & 0.056 & \best{0.290} & 0.275 & \best{0.294} & \best{0.544} & \best{0.396} \\
      \midrule
      \midrule
      \multicolumn{8}{c}{\textit{MIMIC-CXR Findings + Impression}}\\
      \midrule
      MedRegA~\citep{wang2025interpretable} & 0.405 & 0.126 & 0.319 & 0.276 & - & - & - \\
      CXRMate~\citep{NICOLSON2024101585} & - & 0.074 & 0.158 & 0.255 & - & 0.378 & - \\
      MLRG~\citep{Liu2025EnhancedContrastive} & 0.402 & \best{0.152} & 0.172 & \best{0.327} & 0.289 & 0.509 & -\\
      Flamingo-CXR~\citep{Tanno2025collaborationcliniciansvision} & - & 0.101 & - & 0.297 & 0.205 & 0.519 & -\\
      MedGemma~\citep{sellergren2025medgemmatechnicalreport} & 0.333 & 0.052 & 0.338 & 0.197 & 0.209 & 0.523 & 0.388\\
      MARL-Rad (Ours) & \best{0.598} & 0.142 & \best{0.397} & 0.292 & \best{0.316} & \best{0.556} & \best{0.398}\\
      \bottomrule
    \end{tabular}
  }
  \label{tab:mimic_cxr_results}
\end{table*}

%% file: tables/iu_xray_results.tex
\begin{table*}
  \centering
  \setlength{\tabcolsep}{3.6pt}
  \caption{
   Comparison with previous state-of-the-art methods on the IU X-ray dataset~\citep{iu-xray}.
    The best score for each metric within each section is highlighted in \textbf{bold}.
    Our approach achieves the best performance across all clinical efficacy (CE) metrics, which are considered to align more closely with clinicians' assessments~\citep{Tanno2025collaborationcliniciansvision,pmlr-v106-liu19a,pmlr-v116-boag20a,Yu2023EvaluatingProgressRRG}.
  }
  \scalebox{0.85}{
    \begin{tabular}{cccccccc}
      \toprule
      \multirow{2}{*}{\raisebox{-1mm}{\textbf{Method}}}
        & \multicolumn{4}{c}{\textbf{NLG Metrics}$\ \uparrow$}
        & \multicolumn{3}{c}{\textbf{CE Metrics}$\ \uparrow$}\\
      \cmidrule(r){2-5}
      \cmidrule(r){6-8}
      &  \textbf{BLEU-1} & \textbf{BLEU-4}& \textbf{METEOR}& \textbf{ROUGE-L} & \textbf{RadGraph F1} & \textbf{CheXbert F1} & \textbf{GREEN}\\
      \midrule
      \midrule
      \multicolumn{8}{c}{\textit{IU X-ray Findings}}\\
      \midrule
      R2Gen~\citep{chen-etal-2020-generating} & 0.470 & 0.165 & 0.187 & 0.371 & - & - & - \\
      R2GenCMN~\citep{chen-etal-2021-cross-modal} & 0.475 & 0.170 & 0.191 & 0.376 & - & - & - \\
      PPKED~\citep{9578840} & 0.483 & 0.168 & 0.190 & 0.376 & - & - & - \\
      METransformer~\citep{10203079} & 0.483 & 0.172 & 0.192 & 0.380 & - & - & - \\
      KiUT~\citep{10203622} & 0.525 & 0.185 & 0.242 & 0.409 & - & - & - \\
      CoFE~\citep{10.1007/978-3-031-72775-7_10} & - & 0.175 & 0.202 & \best{0.438} & - & - & - \\
      CMCL~\citep{liu-etal-2021-competence} & 0.473 & 0.162 & 0.186 & 0.378 & - & - & - \\
      MAN~\citep{Shen_Pei_Liu_Tian_2024} & 0.501 & 0.170 & 0.213 & 0.386 & - & - & - \\
      CMN~\citep{chen-etal-2021-cross-modal} & 0.475 & 0.170 & 0.191 & 0.375& - & - & - \\
      Med-LMM~\citep{10.1145/3664647.3680760} & - & 0.168 & \best{0.381}  & - & - & - & -\\
      FMVP~\citep{10356722} & 0.485 & 0.169 & 0.201 & 0.398 & - & - & - \\
      LM-RRG~\citep{zhou2024largemodeldrivenradiology} & - & 0.208 & 0.216 & 0.387 & - & - & -\\
      CXRMate~\citep{NICOLSON2024101585} & - & 0.046 & - & 0.282 & 0.291 & 0.277 & -\\
      I3+C2FD~\citep{Liu_Tian_Chen_Song_Zhang_2024} & 0.499 & 0.184 & 0.208 & 0.390 & - & - & - \\
      SRRG~\citep{Hou_2025_ICCV} & 0.533 & 0.218 & 0.219 & 0.418 & - & - & -\\
      MPO~\citep{DBLP:conf/aaai/Xiao0L0B25} & \best{0.548} & 0.209 & 0.224 & 0.415 & - & - & -\\
      DAMPER~\citep{Huang_Chen_Liu_Lu_Luo_Shen_2025} & 0.520 & \best{0.225} & 0.284 & 0.397 & - & - & -\\
      Multi-Agent~\citep{yi2025multimodalmultiagentframeworkradiology} & - & 0.047 & 0.362 & 0.247 & - & - & - \\
      OISA~\citep{xiao-etal-2025-online} & 0.431 & 0.131 & - & - & 0.282 & 0.219 & 0.481\\
      MedGemma~\citep{sellergren2025medgemmatechnicalreport} & 0.179 & 0.012 & 0.287 & 0.215 & 0.270 & 0.427 & 0.632\\
      MARL-Rad (Ours) & 0.468 & 0.046 & 0.347 & 0.292 & \best{0.337} & \best{0.501} & \best{0.644} \\
      \midrule
      \midrule
      \multicolumn{8}{c}{\textit{IU X-ray Findings + Impression}}\\
      \midrule
      CXRMate~\citep{NICOLSON2024101585} & - & 0.046 & - & 0.282 & 0.291 & 0.277 & - \\
      MedGemma~\citep{sellergren2025medgemmatechnicalreport} & 0.244 & 0.058 & 0.432 & 0.212 & 0.265 & 0.469 & 0.653\\
      MARL-Rad (Ours) & \best{0.591} & \best{0.182} & \best{0.531} & \best{0.348} & \best{0.365} & \best{0.516} & \best{0.659} \\
      \bottomrule
    \end{tabular}
  }
  \label{tab:iu_xray_results}
  \vspace{2mm}
\end{table*}

%% file: tables/ablation_study.tex
\begin{table*}
  \centering
  \caption{
    Ablation study results.
    ``MedGemma (vanilla)'' represents the unmodified model without reinforcement learning or agentization.
    ``MedGemma (agent)'' employs the same agentic workflow as MARL-Rad but without any RL optimization.
    ``MedGemma + RL''  applies GSPO to MedGemma in a single-agent setting.
    ``Counterfactual Reward'' uses agent-specific rewards based on counterfactual removal of each regional agent's output from the global agent's context.
    The best and worst scores for each metric within each dataset are highlighted in \textbf{bold} and underline, respectively.
  }
  \vspace{2pt}
  \scalebox{0.8}{
    \begin{tabular}{cccccccc}
      \toprule
      \multirow{2}{*}{\raisebox{-1mm}{\textbf{Method}}}
        & \multicolumn{4}{c}{\textbf{NLG Metrics}$\ \uparrow$}
        & \multicolumn{3}{c}{\textbf{CE Metrics}$\ \uparrow$}\\
      \cmidrule(r){2-5}
      \cmidrule(r){6-8}
      &  \textbf{BLEU-1} & \textbf{BLEU-4}& \textbf{METEOR}& \textbf{ROUGE-L} & \textbf{RadGraph F1} & \textbf{CheXbert F1} & \textbf{GREEN}\\
      \midrule
      \midrule
      \multicolumn{8}{c}{\textit{MIMIC-CXR Findings + Impression}}\\
      \midrule
      MedGemma (vanilla) & 0.333 & 0.052 & 0.338 & 0.197 & 0.209 & 0.523 & 0.388\\
      MedGemma (agent) &  \worst{0.252} & \worst{0.044} & \worst{0.267} & \worst{0.144} & \worst{0.106} & \worst{0.309} & \worst{0.344}\\
      MedGemma + RL & 0.465 & 0.116 & 0.373 & 0.283 & 0.299 & 0.502 & 0.377 \\
      Counterfactual Reward & 0.575 & 0.136 & 0.375 & 0.272 & 0.292 & 0.518 & 0.369 \\
      MARL-Rad (Ours) & \best{0.598} & \best{0.142} & \best{0.397} & \best{0.292} & \best{0.316} & \best{0.556} & \best{0.398} \\
      \midrule
      \midrule
      \multicolumn{8}{c}{\textit{IU X-ray Findings + Impression}}\\
      \midrule
      MedGemma (vanilla) & 0.244 & 0.058 & 0.432 & 0.212 & 0.265 & 0.469 & 0.653\\
      MedGemma (agent) & \worst{0.210} & \worst{0.053} & \worst{0.380} & \worst{0.163} & \worst{0.141} & \worst{0.348} & 0.649 \\
      MedGemma + RL & \best{0.598} & 0.181 & 0.529 & 0.339 & 0.358 & 0.495 & 0.629 \\
      Counterfactual Reward & 0.561 & 0.166 & 0.513 & 0.323 & 0.349 & 0.478 & \worst{0.614} \\
      MARL-Rad (Ours) & 0.591 & \best{0.182} & \best{0.531} & \best{0.348} & \best{0.365} & \best{0.516} & \best{0.659} \\
      \bottomrule
    \end{tabular}
  }
  \label{tab:ablation_study}
\end{table*}

%% file: tables/region_analysis.tex
\begin{wraptable}{r}{0.52\textwidth}
  \vspace{-5.2mm}
  \centering
  \setlength{\tabcolsep}{3.6pt}
  \caption{
    Laterality-specific RadGraph F1 scores.
    ``Left Agent'' and ``Right Agent'' denote the pre-integration outputs of the corresponding regional agents.
    Higher scores indicate better laterality consistency.
    The best score among the final-report methods, i.e., MedGemma w/ RL and MARL-Rad (Ours), is highlighted in \textbf{bold}.
  }
  \scalebox{0.84}{
  \begin{tabular}{ccccc}
    \toprule
    Method
      & \textbf{Left} & \textbf{Right} & \textbf{Left lung} & \textbf{Right lung}\\
    \midrule
    \midrule
    \multicolumn{5}{c}{\textit{MIMIC-CXR Findings + Impression}}\\
    \midrule
    MedGemma w/ RL & 0.144 & 0.142 & 0.219 & 0.102 \\
    MARL-Rad (Ours) & \best{0.184} & \best{0.277} & \best{0.406} & \best{0.146}\\
    \midrule
    Left Agent & 0.214 & 0.011 & 0.369 & 0.066 \\
    Right Agent & 0.001 & 0.257 & 0.000 & 0.156 \\
    \midrule
    \midrule
    \multicolumn{5}{c}{\textit{IU X-ray Findings + Impression}}\\
    \midrule
    MedGemma w/ RL & 0.063 & 0.044 & \best{0.035} & 0.057 \\
    MARL-Rad (Ours) & \best{0.082} & \best{0.166} & 0.032 & \best{0.168}\\
    \midrule
    Left Agent & 0.130  & 0.005 & 0.155 & 0.082 \\
    Right Agent & 0.000 & 0.169 & 0.000 & 0.175 \\
    \bottomrule
  \end{tabular}
  }
  \label{tab:region_analysis}
  \vspace{-2mm}
\end{wraptable}

%% file: figures/case_study.tex
\newcommand{\caseStudyFont}{\fontsize{6pt}{6pt}\selectfont\sffamily}
\newcommand{\caseNoteFont}[1]{\fontsize{6pt}{6pt}\selectfont\textbf{\textsf{\color{myBlue}#1}}}
\begin{figure}[t]
  \centering
  \begin{tikzpicture}[scale=0.8, every node/.style={scale=0.8}]
    \filldraw[fill=myYellow!30, draw=myYellow!80] (0mm, 10mm) rectangle ++(84mm, -19mm);
    \filldraw[fill=myPink!30, draw=myPink!80] (0mm, -10mm) rectangle ++(84mm, -17mm);
    \filldraw[fill=myOrange!30, draw=myOrange!80] (0mm, -28mm) rectangle ++(84mm, -16.5mm);

    \draw[->, myBlue, very thick] (85mm, 0.5mm) -- ++(4mm, 0);
    \draw[-, myBlue, very thick] (85mm, -18.5mm) -- ++(2mm, 0);
    \draw[-, myBlue, very thick] (85mm, -36.25mm) -- ++(2mm, 0) -- ++(0, 36.75mm);

    \filldraw[fill=myBlue!30, draw=myBlue!80] (90mm, 10mm) rectangle ++(84mm, -19mm);

    \filldraw[fill=myGreen!30, draw=myGreen!80] (90mm, -10mm) rectangle ++(84mm, -23.5mm);

    \node at (160mm, -35mm) {\includegraphics[width=20mm]{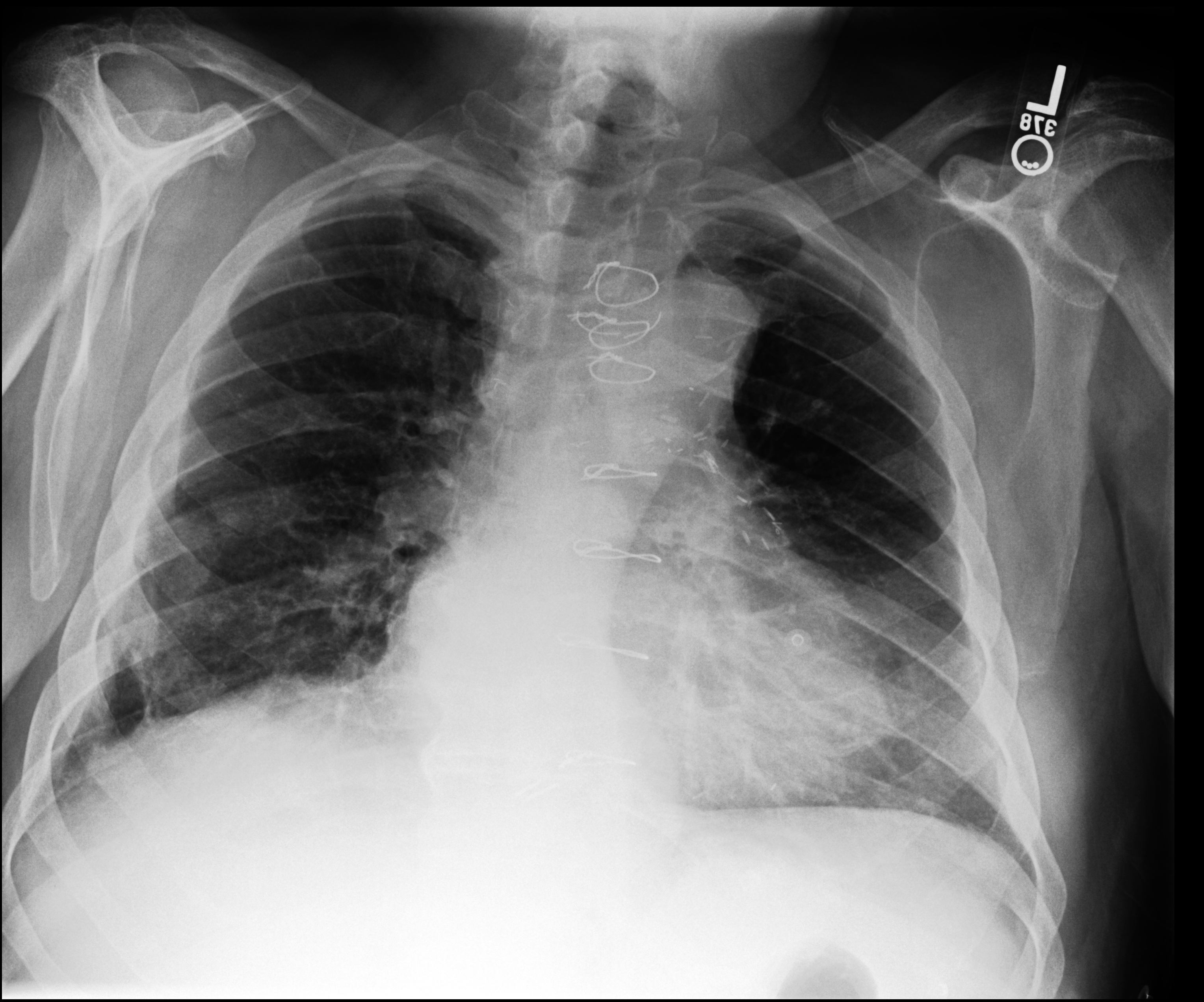}};
    \node at (160mm, -45mm) {
      \caseNoteFont{Input Chest X-ray Image}
    };

    \node[anchor=east] at (84mm, 8.5mm) {
      \caseNoteFont{Output of Left-Region Agent}
    };
    \node[anchor=west, align=left] at (0mm, 0.5mm) {
      \parbox{81mm}{\caseStudyFont
        \#\#\# Findings\\
        The left lung is relatively clear. The left hilar structures are unremarkable. The left costophrenic angle is sharp. The left clavicle is intact. The left hemithorax is unremarkable. The left apical zone is unremarkable.\\
        \#\#\# Impression\\
        There is no evidence of acute infiltrate in the left lung. The left hilar structures are unremarkable. The left costophrenic angle is sharp. The left clavicle is intact. The left hemithorax is unremarkable. The left apical zone is unremarkable.
      }
    };
    \node[anchor=east] at (84mm, -11.5mm) {
      \caseNoteFont{Output of Central-Region Agent}
    };
    \node[anchor=west, align=left] at (0mm, -18.5mm) {
      \parbox{80mm}{\caseStudyFont
        \#\#\# Findings\\
        The heart size is normal. The mediastinum is unremarkable. The aortic arch is visualized. The trachea and carina are midline. The SVC is visualized. The atria and cavoatrial junction are visualized. The abdomen is not well visualized. The spine shows degenerative changes. No acute findings in the lungs.\\
        \#\#\# Impression\\
        No acute findings.
      }
    };
    \node[anchor=east] at (84mm, -29.5mm) {
      \caseNoteFont{Output of Right-Region Agent}
    };
    \node[anchor=west, align=left] at (0mm, -36.25mm) {
      \parbox{80mm}{\caseStudyFont
        \#\#\# Findings\\
        The right lung is clear. The right hilar structures are unremarkable. The right costophrenic angle is sharp. The right hemidiaphragm is visible. The right clavicle is visible. The right apical zone is visible.\\
        \#\#\# Impression\\
        The right lung is clear. There is no evidence of infiltrate, pneumothorax, or pleural effusion.
      }
    };

    \node[anchor=east] at (174mm, 8.5mm) {
      \caseNoteFont{Output of Global-Integrating Agent}
    };
    \node[anchor=west, align=left] at (90mm, 0.5mm) {
      \parbox{80mm}{\caseStudyFont
        \#\#\# Findings\\
        The lungs are clear without focal consolidation, pleural effusion, or pneumothorax. The heart size is normal. The mediastinal contours are unremarkable. There is evidence of prior median sternotomy with surgical clips noted in the anterior mediastinum. There is mild emphysema, particularly in the lung apices. There is also evidence of right pleural thickening.\\
        \#\#\# Impression\\
        No acute cardiopulmonary process.
      }
    };

    \node[anchor=east] at (174mm, -11.5mm) {
      \caseNoteFont{Ground Truth Report}
    };
    \node[anchor=west, align=left] at (90mm, -21.75mm) {
      \parbox{80mm}{\caseStudyFont
        \#\#\# Findings\\
        The cardiomediastinal and hilar contours are stable. The aorta is again noted to be tortuous. The patient is status post CABG with median sternotomy wires in place. The second most superior median sternotomy wires again noted to be fractured. There is no pleural effusion or pneumothorax. The lungs are well-expanded with stable scarring at the right costophrenic angle. There is no new focal consolidation concerning for pneumonia. There is no overt pulmonary edema. The upper abdomen is unremarkable aside from surgical clips.\\
        \#\#\# Impression\\
        No acute cardiopulmonary process.
      }
    };

  \end{tikzpicture}
  \vspace{-5mm}
  \caption{
    Example output from MARL-Rad.
    Region-specific agents focus on their assigned regions and generate regional diagnoses. The global integrating agent synthesizes these drafts into a concise and coherent report while adding relevant global findings.
    The resulting final report is detailed and aligns well with the ground truth, correctly concluding with ``No acute cardiopulmonary process.''
  }
  \label{fig:case_study}
\end{figure}

%% file: figures/comparison_case.tex
\begin{wrapfigure}{r}{0.45\textwidth}
  \vspace{-8mm}
  \centering
  \begin{tikzpicture}[scale=0.8, every node/.style={scale=0.8}]
    \filldraw[fill=black!10, draw=black!15] (0mm, 10mm) rectangle ++(78mm, -19mm);
    \node[anchor=west, align=left] at (0mm, 0.5mm) {
      \parbox{76mm}{\fontsize{8pt}{8pt}\selectfont\sffamily
        \#\#\# Findings\\
        The lungs are clear. There is no focal consolidation, pleural effusion, or pneumothorax. The heart size is normal. The mediastinal and hilar contours are unremarkable.\\
        \#\#\# Impression\\
        No acute cardiopulmonary process.
      }
    };
  \end{tikzpicture}
  \vspace{-6.4mm}
  \caption{
    An example case generated by the single-agent RL model (MedGemma w/ RL) using the same input image as in \cref{fig:case_study}.
  }
  \label{fig:case_study_comparison}
\vspace{-3mm}
\end{wrapfigure}

%% file: sec/human_evaluation.tex
\section{Clinician evaluation}
\label{sec:clinician_evaluation}

\input{figures/human_evaluation}

Automated metrics may not fully capture clinical quality, and improvements on such metrics may not always align with clinician judgment.
To partially address this concern, we conduct a small blinded clinician comparison between MARL-Rad outputs and ground-truth reports.

To minimize bias, the two reports for each case were anonymized and randomly ordered.
Physicians evaluated each pair together with the corresponding chest X-ray image, without knowing the source of either report.
For each case, physicians rated five aspects: completeness, correctness, conciseness, readability, and clinical utility.
Detailed definitions of these evaluation criteria are provided in \cref{appx:clinician_evaluation}.
Each aspect was evaluated on a 5-point comparative Likert scale with the options ``Report A is better,'' ``Report A is slightly better,'' ``About the same,'' ``Report B is slightly better,'' and ``Report B is better.''
In addition, physicians provided an overall preference vote between reports A and B.
We randomly sampled 10 images from the MIMIC-CXR test set, and six physicians completed the evaluation.
The results are shown in \cref{fig:physician_preferences}.

Across all five criteria, physician ratings show that MARL-Rad outputs are clinically comparable to the ground-truth reports.
The overall preference vote is nearly balanced, with a slight preference for MARL-Rad.
These results complement the automated evaluation and suggest that the gains of MARL-Rad do not merely reflect overfitting to automatic metrics, but correspond to clinically comparable report quality.

%% file: figures/human_evaluation.tex
\begin{figure}[t]
  \centering
  \hspace*{-13mm}\includegraphics[width=1.12\textwidth]{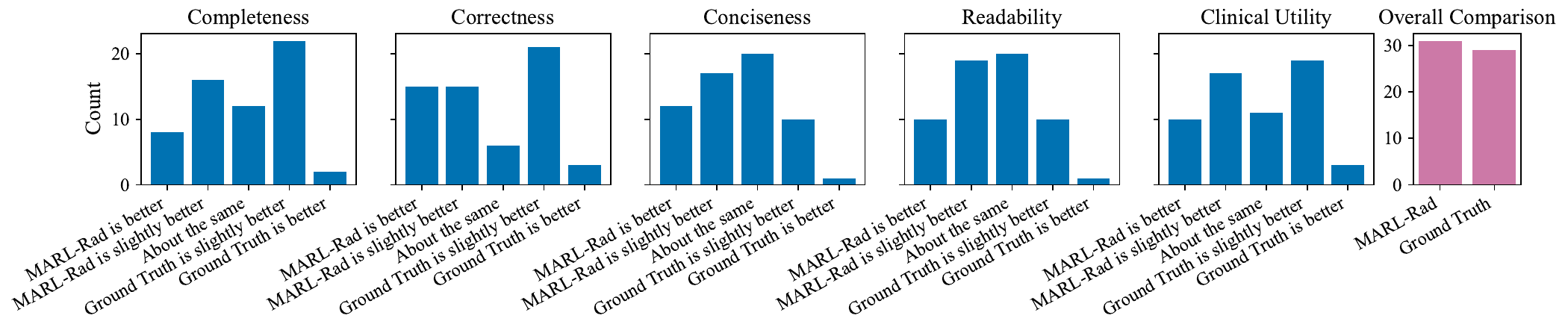}
  \vspace{-5mm}
  \caption{
    Physician preferences comparing MARL-Rad and the ground-truth reports.
    Bars show 5-point Likert ratings for five criteria; the rightmost panel shows the overall 2-choice vote.
    During evaluation, report sources were anonymized and randomly shuffled for each case.
  }
  \label{fig:physician_preferences}
  \vspace{1mm}
\end{figure}

%% file: sec/conclusion.tex
\section{Conclusion}
\label{sec:conclusion}

In this work, we introduced MARL-Rad, a novel multi-modal multi-agent reinforcement learning framework for radiology report generation.
MARL-Rad coordinates region-specific agents with a global integrating agent and jointly optimizes the entire agent system end-to-end to produce clinically consistent reports.
By optimizing agents on policy within their deployed workflow, MARL-Rad overcomes the limitations of training-free agentification of fixed LLMs.
Our method achieves state-of-the-art performance on CE metrics on both MIMIC-CXR and IU X-ray datasets, enhances laterality consistency, and yields more accurate, detailed reports.
Although this work focuses on CXR-based RRG, the framework may be extended to other workflow-structured applications.
Future work includes applying this framework to other modalities in medical AI, such as electrocardiogram (ECG) and echocardiography, as well as other real-world tasks.

%% file: apx/extended_related_work.tex
\section{Extended related work}
\label{apx:extended-related-work}

\subsection{Reinforcement learning for LLMs}

Reinforcement Learning with Verifiable Rewards (RLVR) has recently emerged as an alternative to traditional Reinforcement Learning from Human Feedback (RLHF), providing objective, outcome-based feedback through deterministic verification functions rather than human preference models.
By rewarding correctness or rule-based validity, RLVR has improved reasoning capabilities in structured domains, such as mathematics and code generation~\citep{%
  wen2025reinforcementlearningverifiablerewards,
  su2025crossingrewardbridgeexpanding,
  shao2024deepseekmathpushinglimitsmathematical,
  yu2025dapoopensourcellmreinforcement,
  zheng2025groupsequencepolicyoptimization,
  liu2025understanding,
  liu2025itrickstrapsdeep
}.
More recently, agentic reinforcement learning has gained attention, where models interact with external tools such as Python interpreters or web search engines to improve factual accuracy~\citep{singh2025agenticreasoningtoolintegration,mai2025agentrlscalinglaw,dong2025toolstarempoweringllmbrainedmultitool,zhang2025l0reinforcementlearninggeneral,feng2025retoolreinforcementlearningstrategic,qian2025toolrlrewardtoollearning,li2025torlscalingtoolintegratedrl,song2025r1searcherincentivizingsearchcapability,chen2025research}.
However, most of these methods still focus on training a single, monolithic model, rather than jointly optimizing multiple agents. Consequently, genuine cooperative optimization among agents in real-world workflows remains largely unexplored.

\subsection{Agentic systems with LLMs}

Agentic systems leverage LLMs to perform goal-oriented reasoning, planning, and collaboration through structured interactions among multiple agents.
Recent frameworks such as AutoGen~\citep{wu2024autogen}, MetaGPT~\citep{hong2024metagpt}, CAMEL~\citep{li2023camel}, have shown that pretrained LLMs can engage in cooperative behaviors via carefully designed prompts and workflows~\citep{%
  hu2025owl,%
  baba2025proveragentagentbasedframework,%
  zhao2025llmbasedagenticreasoningframeworks,%
  plaat2025agenticlargelanguagemodels,%
  koubaa2025pretrainedlanguageagentic,%
  jiang2025largeaimodelsagentic,%
  Li2024surveyllmbasedmultiagentsystems,%
  liang-etal-2024-encouraging%
}.
However, most of these systems are training-free, relying on pre-trained models without end-to-end optimization, resulting in suboptimal coordination and limited adaptability when applied to complex, real-world workflows.

Recent works have started to explore reinforcement learning within agentic systems.
For example, MALT~\citep{motwani2025malt} performs post-training using trajectories collected from agent executions, but its optimization remains off-policy and detached from real-world workflow interactions.
MAPoRL~\citep{park-etal-2025-maporl} applies multi-agent reinforcement learning to improve collaboration among language models, yet the optimization of realistic, role-structured workflows in applied domains remains largely unexplored.
To address this gap, AgentFlow~\citep{li2025intheflowagenticoptimizationeffective} introduces on-policy reinforcement learning within an actual multi-agent workflow, but the optimization is limited to a single key planner agent rather than the entire agentic system.

%% file: apx/experimental_setup.tex
\section{Detailed experimental setup}
\label[appendix]{appx:experimental_setup}

\cref{sec:experimental_setup} provides an overview of the experimental setup. In this section, we describe additional details that were not included in \cref{sec:experimental_setup}.

\paragraph{Datasets}
MIMIC-CXR~\citep{mimic-cxr} is a large-scale publicly available chest X-ray dataset containing more than 370,000 images paired with free-text radiology reports. The dataset covers a wide range of thoracic conditions and includes both frontal and lateral views.
IU X-ray (Indiana University Chest X-Ray dataset)~\citep{iu-xray} consists of chest X-ray images paired with corresponding radiology reports, including both Findings and Impression sections. Each study contains frontal and lateral views with detailed narrative annotations.

\paragraph{Implementation details}
The reinforcement learning implementation is built on top of verl~\citep{10.1145/3689031.3696075}, using vLLM~\citep{10.1145/3600006.3613165} for rollouts and Fully Sharded Data Parallel (FSDP)~\citep{10.14778/3611540.3611569} for parameter updates.
We set the batch size to 16, and 16 rollouts are generated for each training sample.
The maximum rollout length is set to 2048 tokens.
We train each model for 100 reinforcement learning update steps.
Following the original GSPO paper~\citep{zheng2025groupsequencepolicyoptimization}, we set the clipping thresholds in objectives to $\varepsilon_\text{high}=0.0004$ and $\varepsilon_\text{low}=0.0003$. The rollout temperature is fixed at 1.0. We use a learning rate of 1e-6, a warmup ratio of 0.05, and a weight decay of 0.1. We use AdamW as the optimizer. The batch size for parameter updates is set to 4.
We use \texttt{google/medgemma-4b-it}\footnote{\url{https://huggingface.co/google/medgemma-4b-it}} as the base checkpoint.
We use verl v0.5.0, vLLM v0.10.1, Transformers v4.53.3, and PyTorch v2.7.1.
All experiments are conducted on 4 $\times$ NVIDIA H100 GPUs, each with 80 GB of memory.
Each training run took approximately 18 hours.
For inference, we follow the default vLLM generation settings except that the maximum number of generated tokens is set to 2048.

\section{Computational cost and deployment considerations}
\label{appx:computational_cost}

To assess the practical deployment cost of MARL-Rad, we measured inference latency and GPU memory usage on H100 GPUs using 600 reports from the MIMIC-CXR test set.
When the three region-specific agents were executed in parallel, MARL-Rad required 772 seconds in total, corresponding to 1.28 seconds per sample.
When all agents were executed sequentially, the total inference time was 1586 seconds, corresponding to 2.64 seconds per sample.
This latency is identical to that of the training-free agentic baseline used in our ablation study, i.e., ``MedGemma (agent),'' because both methods share the same multi-agent architecture at inference time.
For reference, the non-agent baseline, ``MedGemma (vanilla),'' required 398 seconds in total, corresponding to 0.66 seconds per sample.
The GPU memory usage was approximately 9 GB per agent.

Although MARL-Rad introduces additional computational overhead due to agentization, the region-specific agents can be executed in parallel, substantially reducing latency compared with sequential execution.
The resulting inference speed remains within a practical range for real-world radiology reporting workflows.

\section{Details of the laterality-specific RadGraph F1 score computation}
\label[appendix]{appx:detail_region_analysis}

To evaluate region-level consistency focused on laterality, we compute a laterality-specific RadGraph F1 score derived from the standard RadGraph evaluation.
We first construct the RadGraph for both the prediction and the ground truth using the standard RadGraph extraction procedure. From each constructed graph, we then extract the entities corresponding to laterality-related anatomical regions, which include the tokens ``left'', ``right'', ``left lung'', and ``right lung.''
From these entities, we further extract the subgraph consisting of all nodes and relations connected to them, resulting in a laterality-specific subset of each RadGraph.
We then calculate the laterality-specific RadGraph F1 score by applying the standard RadGraph matching procedure but restrict the comparison to this laterality-specific subgraph only.

\section{Clinician evaluation criteria}
\label{appx:clinician_evaluation}

Physicians evaluated each report pair according to the following five criteria.

\begin{itemize}
  \item \textbf{Completeness}: Whether the report includes all clinically relevant findings visible in the chest X-ray, without omitting important abnormalities or normal findings that should be documented.

  \item \textbf{Correctness}: Whether the described findings are clinically accurate and consistent with the chest X-ray, without introducing incorrect diagnoses, false findings, or contradictions.

  \item \textbf{Conciseness}: Whether the report is appropriately concise, avoiding unnecessary repetition, irrelevant descriptions, or overly verbose statements while preserving clinically important information.

  \item \textbf{Readability}: Whether the report is clearly written, well organized, and easy for clinicians to understand, with coherent phrasing and appropriate radiological terminology.

  \item \textbf{Clinical Utility}: Whether the report would be useful in clinical practice for supporting diagnosis, communication, and downstream decision-making.
\end{itemize}

Physicians were instructed to compare two anonymized reports for each chest X-ray case and select which report was better for each criterion.
The report order was randomized, and the physicians were blinded to whether each report was generated by MARL-Rad or taken from the ground truth.
The evaluation was conducted with approval from the affiliated hospital according to the applicable institutional requirements.
Potential risks to participants were minimal, as the task involved expert assessment of anonymized reports without identifiable patient information.
Physicians performed the evaluation during regular working hours under the affiliated hospital's approval, and no additional honorarium was provided specifically for this task.

\section{Limitations and broader impact}
\label{sec:limitations}

\subsection{Limitations}

Although MARL-Rad achieves strong performance on standard RRG benchmarks, this work has several limitations.
First, our experiments focus on chest X-ray report generation using MIMIC-CXR and IU X-ray, and the effectiveness of the proposed framework on other imaging modalities, institutions, and clinical settings remains to be validated.
Second, our reinforcement learning rewards rely on automatic clinical metrics such as CheXbert and RadGraph.
While these metrics are clinically motivated and widely used, they may not fully capture all aspects of report quality and may contain annotation or evaluation noise.
We partially address this concern through blinded clinician evaluation, but the evaluation is limited in scale and should be expanded in future work.
Third, MARL-Rad requires multiple model invocations at inference time, increasing computational cost and GPU memory usage compared with a single-model baseline, although parallel execution mitigates the added latency as discussed in \cref{appx:computational_cost}.
Finally, although our framework represents a step toward clinically useful agentic RRG systems by improving clinical efficacy metrics and laterality consistency, real-world deployment would require rigorous prospective validation, safety monitoring, and integration into radiologist-supervised workflows.

\subsection{Broader impact}

MARL-Rad aims to improve radiology report generation by training agentic systems within the workflows in which they are deployed.
If properly validated, such systems could help reduce the burden of report writing, improve the consistency of generated reports, and support radiologists in clinical documentation.

At the same time, medical report generation is a high-stakes application.
Incorrect or hallucinated findings could negatively affect clinical decision-making if used without appropriate oversight.
Future deployment should require careful clinical validation, bias assessment across patient populations and institutions, privacy-preserving data handling, and mechanisms for human review and correction.